\documentclass[journal]{IEEEtran}
\usepackage{graphicx}
\usepackage{subfigure}
\usepackage{cite}
\usepackage{amsmath,amssymb} 
\usepackage[colorlinks=false]{hyperref}
\usepackage{color}
\usepackage[ruled]{algorithm2e}

\newcommand{\tabincell}[2]{\begin{tabular}{@{}#1@{}}#2\end{tabular}}

\begin{document}
%
\title{Network Pruning using Adaptive Exemplar Filters}
%
%
%

\author{Mingbao Lin, Rongrong Ji, Shaojie Li, Yan Wang
\thanks{This work was supported in part by the Nature Science Foundation of China under Grant U1705262}
\thanks{J. Doe and J. Doe are with Anonymous University.}
\thanks{Manuscript received April 19, 2005; revised August 26, 2015.}}

\author{Mingbao Lin,
        Rongrong Ji,~\IEEEmembership{Senior Member,~IEEE},
        Shaojie Li, 
        Yan Wang, \\
        Yongjian Wu, 
        Feiyue Huang,
        Qixiang Ye,~\IEEEmembership{Senior Member,~IEEE}
\IEEEcompsocitemizethanks{
\IEEEcompsocthanksitem M. Lin, R. Ji (Corresponding Author) and S. Li are with the Media Analytics and Computing Laboratory, Department of Artificial Intelligence, School of Informatics, Xiamen University, Xiamen 361005, China (e-mail: rrji@xmu.edu.cn).\protect
\IEEEcompsocthanksitem R. Ji is also with Institute of Artificial Intelligence, Xiamen University.\protect
\IEEEcompsocthanksitem Y. Wang is with Pinterest, USA.\protect
\IEEEcompsocthanksitem Y. Wu and F. Huang are with Youtu Laboratory, Tencent, Shanghai 200233, China.\protect
\IEEEcompsocthanksitem Q. Ye is with School of Electronic, Electrical and Communication Engineering, University of Chinese Academy of Sciences, Beijing 101408, China.\protect 
\IEEEcompsocthanksitem R. Ji and Q. Ye are also with Peng Cheng Lab, Shenzhen, China.\protect 
}
\thanks{Manuscript received April 19, 2005; revised August 26, 2015.}}

\markboth{IEEE TRANSACTIONS ON NEURAL NETWORKS AND LEARNING SYSTEMS}%
{Shell \MakeLowercase{\textit{et al.}}: Bare Demo of IEEEtran.cls for IEEE Journals}

\maketitle

\begin{abstract}
Popular network pruning algorithms reduce redundant information by optimizing hand-crafted models, and may cause suboptimal performance and long time in selecting filters. We innovatively introduce adaptive exemplar filters to simplify the algorithm design, resulting in an automatic and efficient pruning approach called EPruner.
Inspired by the face recognition community, we use a message passing algorithm Affinity Propagation on the weight matrices to obtain an adaptive number of exemplars, which then act as the preserved filters. EPruner breaks the dependency on the training data in determining the ``important'' filters and allows the CPU implementation in seconds, an order of magnitude faster than GPU based SOTAs.
Moreover, we show that the weights of exemplars provide a better initialization for the fine-tuning. On VGGNet-16, EPruner achieves a 76.34\%-FLOPs reduction by removing 88.80\% parameters, with 0.06\% accuracy improvement on CIFAR-10. In ResNet-152, EPruner achieves a 65.12\%-FLOPs reduction by removing 64.18\% parameters, with only 0.71\% top-5 accuracy loss on ILSVRC-2012. Our code can be available at \url{https://github.com/lmbxmu/EPruner}.


%
\end{abstract}

\section{Introduction}\label{introduction}
\IEEEPARstart{C}{onvolutional} Neural Networks (CNNs) and their variants have been applied to a broad range of problems such as image classification \cite{simonyan2015very}, action recognition \cite{chen2017deep}, semantic segmentation \cite{long2015fully} and few-shot learning~\cite{peng2019few}. However, most high-performing CNNs are designed to execute on high-end GPUs with substantial memory and computational power, which hinders their practical applications in resource-constrained environments, such as mobile and embedded devices. Therefore, model compression techniques, \emph{e.g.}, low-rank decomposition \cite{denton2014exploiting,hayashi2019exploring,kim2019efficient}, parameter quantization \cite{cheng2017quantized,krishnamoorthi2018quantizing,wang2019haq}, and network pruning \cite{wang2017novel,he2018amc,chen2018constraint,frankle2019lottery}, have been proposed. Among them, network pruning has been widely studied and proved to be an effective tool to reduce the network complexity.
In contrast to unstructured weight pruning that removes individual entries in the weight matrices~\cite{lecun1990optimal,guo2016dynamic,frankle2019lottery,ding2019global} and thus requires specialized hardware or software, structured filter pruning has attracted more attention since it removes the entire filters and corresponding channels, which requires no extra requirements for the inference platforms~\cite{hu2016network,he2017channel,liu2017learning,yu2018nisp,zhao2019variational,lin2019towards,liu2019metapruning,lin2020hrank}. 

Based on the filter preserving policy, we empirically classify the filter pruning approaches into three categories: (1) Rule-based pruning decides the pruned network architecture by hand-crafted designation and the pruned network usually inherits the most important filter weights either measured by an intrinsic property of the pre-trained model, \emph{e.g.}, the $\ell_1$-norm \cite{li2017pruning}, or through iterative refinement on training data~\cite{he2017channel,luo2017thinet}. Typically, the designated architecture is sub-optimal. Moreover, it usually performs layer-wise fine-tuning/optimization to recover the accuracy, which is computationally intensive. (2) Regularization-driven pruning retrains networks with hand-crafted constraints, such as sparsity \cite{liu2017learning,lin2019towards,lin2019toward} and budget awareness \cite{lemaire2019structured}. The trained filters below a given threshold are removed and the weights of preserved filters are inherited by the pruned network. However, retraining the model is expensive, and the introduced hyper-parameters also require manual analysis. (3) Architecture-search-driven pruning focuses on searching for a better architecture, typically through heuristic-based policies, such as evolutionary algorithm \cite{liu2019metapruning} or the artificial bee colony algorithm \cite{lin2020channel}. The filter weights of these methods are randomly sampled for the follow-up fine-tuning. However, the architecture search is data-dependent thus also computationally intensive, as validated in Tab.\,\ref{efficiency} of the experiments. Besides, the search results are not deterministic.
Overall, existing methods still heavily depend on hand-crafted rules from humans, and/or involve time-consuming retraining/search.
How to achieve automatic and efficient model compression still remains an open problem.

Thus, existing methods suffer either great cost on human labor or large time complexity in pruning. In this paper, we propose a novel filter pruning method, termed EPruner, which selects the most important filters (exemplars), to solve the above problems. We argue that the removal of informational redundancy is an underlying theme behind all these three types of approaches. As stressed in network interpretation~\cite{li2015convergent,morcos2018importance,zhou2018revisiting}, the filters, even within the same layer, have different impacts. This indicates that different filters have different properties and there exist some exemplars which contribute more in the network inference. Thus, the challenge lies in figuring out the exemplar number without human involvement and identifying which filter could be the exemplar such that the efficient end-to-end fine-tuning can be conducted. Inspired by the face recognition community, we use a graph message-passing algorithm Affinity Propagation~\cite{frey2007clustering} to make our approach adaptive to the pre-trained CNN in both finding the number of exemplars and identifying more informative exemplars.

\begin{figure*}[!t]
\begin{center}
\includegraphics[height=0.33\linewidth]{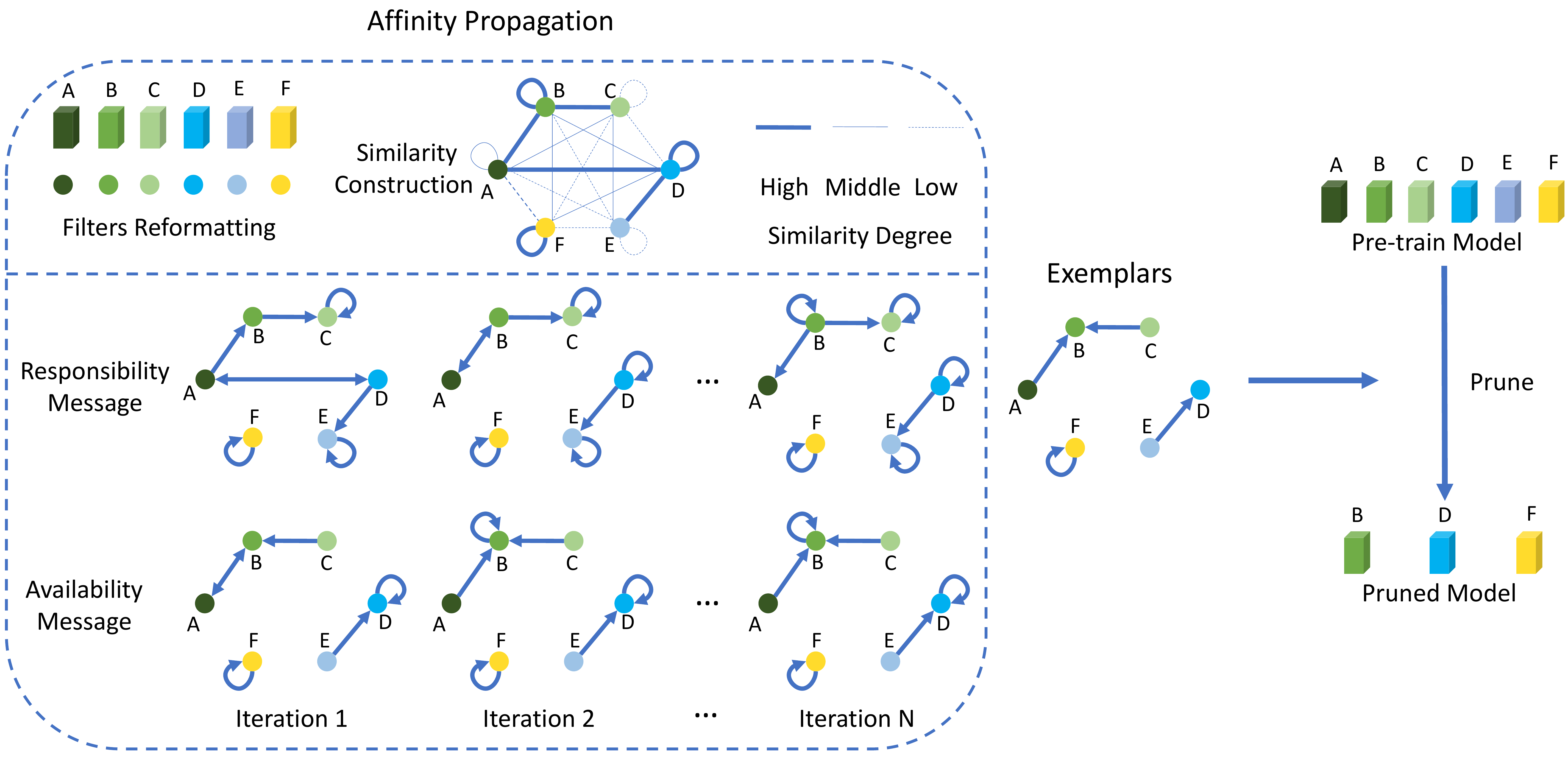}
\end{center}
\caption{\label{framework}
Framework of EPruner. The similarity graph among filters is constructed. The ``responsibility'' r(A, B) indicates the suitedness of filter B to the exemplar of filter A. The ``availability'' a(A, B) reflects the appropriatedness of filter A choosing filter B as its exemplar. The two messages are passing among filters until exemplars merges, which then serve as the preserved filters. More details can be found in Sec.\,\ref{exemplar}.} 
\end{figure*}

%
Specifically, as shown in Fig.\,\ref{framework}, we treat each filter as a data point, and connect them with weighted edges.
Via message passing among the filters, Affinity Propagation could select the representative \emph{exemplars} from all the filters.
From a network interpretation perspective~\cite{li2015convergent,morcos2018importance,zhou2018revisiting}, this can be treated as selecting the most impacting filters from a filter pool.
While the weight matrices in different layers may have different numbers and variances, Affinity Propagation is able to obtain an adaptive number of exemplars, with a non-learning hyper-parameter controlling the compression strength.
To model the similarity between each pair of filters, it is critical to choose a proper distance metric.
However, we found the proposed approach works surprisingly well even with the naive negative Euclidean distance, over-performing state-of-the-art pruning algorithms~\cite{luo2017thinet,he2017channel,huang2018data,lin2019towards,liu2019metapruning,lin2020channel,lin2020hrank} with less time consumption in determining the important filters.
We argue that the innovative introduction of adaptive exemplar filters greatly simplifies the problem structure.
Another counter-intuitive finding is related with the observation~\cite{liu2019rethinking} that randomly initialized weights can perform better than inheriting the most important filters. 
In this paper, we find that proper filter weight initialization based on the exemplars could significantly outperform randomly initialized filter weights, which thus breaks the observation in~\cite{liu2019rethinking}.

As a summary, our contributions are three-fold:
\begin{itemize}
\item Our EPruner is able to figure out an adaptive number of exemplars and identify more informative exemplars in a unified framework. Thus, the network pruning can be efficiently implemented via an end-to-end fine-tuning.
\item With a higher reduction of model complexity, EPruner lessens the involvement of human labor in the pruning. Besides, it also leads to less time consumption in important filter selection and deterministic pruning results.
\item We also provide extensive experiments on VGGNet \cite{simonyan2015very}, GoogLeNet \cite{szegedy2015going} and ResNets \cite{he2016deep} to demonstrate the efficacy of EPruner in reducing the model complexity and better performance in comparison with several state of the arts.
\end{itemize}

In the following, we will first discuss the related work in Sec.\,\ref{related}. Then, we elaborate our EPruner for network pruning in Sec.\,\ref{methodology}. In Sec.\,\ref{experiment}, the experimental results are provided and analyzed. Lastly, we conclude this paper in Sec.\,\ref{conclusion}.

\section{Related Work}\label{related}
Unstructured weight pruning removes the individual neurons in the filter or the connection between fully-connected layers. The pioneering work \cite{lecun1990optimal} utilizes the second derivatives to balance the training loss and model complexity. Han \emph{et al}. \cite{han2015learning} proposed to recursively prune the low-weight connections and retrain the remaining subnetwork with the $\ell_2$ normalization. Dynamic network surgery \cite{guo2016dynamic} performs weight pruning and splicing. The pruning is made to compress the network model and the slicing is used to enable connection recovery. Aghasi \emph{et al}. \cite{aghasi2017net} removed connections by solving a convex optimization program. In \cite{chen2018constraint}, the compression is formulated as the constrained Bayesian optimization solved by the annealing strategy. The lottery ticket hypothesis \cite{frankle2019lottery} randomly initializes a dense network and trains it from scratch. The subnet with high-weight values are extracted, and retrained with the initial weight values of the dense model.

Unstructured weight pruning results in an irregular sparsity, which requires customized hardware and software to support the practical speedup. In contrast, structured filter pruning is especially advantageous by removing the entire filters and the corresponding channels in the next layer directly. It has no extra requirements for the inference engine and thus can be easily deployed.

To this end, rule-based pruning requires human experts to specify the pruned architecture. Hu \emph{et al}. \cite{hu2016network} observed the large output of zero activations in the network and the filter weights with a lower percentage of zero outputs are inherited. The $\ell_1$-norm pruning \cite{li2017pruning} assumes filters with large norms are more informative, and thus the corresponding weights are inherited. Lin \emph{et al}. \cite{lin2020hrank} preserved the filter weights with high-rank feature map outputs. Another direction in filter pruning formulates the pruning as an optimization problem using the statistics information from the next layer \cite{he2017channel} or a linear least square to reconstruct the outputs \cite{luo2017thinet}, and the optimized weights are forwarded for fine-tuning.

Regularization-driven pruning performs model complexity reduction by retraining the network with hand-crafted constraints. Liu \emph{et al}.~\cite{liu2017learning} and Zhao~\emph{et al}. \cite{zhao2019variational} imposed a sparsity constraint on the scaling factor of the batch normalization layer and the weights with higher scaling factors are preserved for the fine-tuning. Huang \emph{et al}. \cite{huang2018data} and Lin \emph{et al}. \cite{lin2019towards} introduced a sparse soft mask on the feature map outputs or the filters. The filters with non-zero masks are preserved and the corresponding trained weights are preserved. Lemaire \emph{et al}. \cite{lemaire2019structured} proposed a knowledge distillation loss function combined with a budget-constrained sparsity loss to train the network and guarantee the neuron budget at the same time.

Architecture-search-driven pruning tries to search for the optimal pruned architecture. In \cite{liu2019metapruning}, a large auxiliary PruningNet has to be trained in advance to evaluate the performance of each potential pruned architecture derived from the evolutionary procedure. In \cite{lin2020channel}, artificial bee colony algorithm is applied to search for the pruned architecture and the accuracy is regarded as the fitness of each architecture. In light of the recent work \cite{liu2019rethinking} which claims that the essence of network pruning lies in finding the optimal pruned architecture rather than selecting the most important filter weights, the initializations of these methods are randomly sampled from the Gaussian distribution \cite{liu2019metapruning} or the pre-trained model \cite{lin2020channel} for fine-tuning. 

Compared with the rule-based pruning, the novelty of our EPruner is two-fold: First, the pruned network architecture and filter selection are automatic and adaptive, which lessens the human involvement. Second, the fine-tuning is implemented in an efficient end-to-end manner rather than the inefficient layer-wise manner. Compared with regularization-driven pruning, the novelty of our EPruner lies in no hand-crafted constraints and training from scratch. Compared with the architecture-search-driven pruning, the novelty of our Epruner falls into its adaptivity to finding optimal pruned architecture without computationally intensive search progress and the pruning result is deterministic.

\section{Methodology}\label{methodology}

\subsection{Problem Definition}\label{preliminary}
Given a pre-trained CNN model $\mathbf{F}$ with a total of $L$ convolutional layers, its filter weights can be represented as a set of 4-D tensors $\mathbf{W} = \{ \mathbf{w}_k \}_{k=1}^L$. The $k$-th layer parameters of $\mathbf{w}_k$ have the shape of $c_{k} \times c_{k-1} \times h_k \times w_k$, where $c_{k}$ represents the number of filters, $c_{k-1}$ represents the number of input channels of each filter, and $h_k$ and $w_k$ represent the height and width of each filter. As can be seen, the number of channels, \emph{i.e.}, $c_{k-1}$, is equal to the number of filters in the $(k-1)$-th layer.

\begin{table}[!t]
\centering
\caption{Notations and Their Meanings.}
\label{notation}
\begin{tabular}{|c|c|}
\hline
Notation            &Meaning     \\
\hline
$\mathbf{w}_k$      &Filters in the $k$-th layer \\\hline

$\mathbf{w}_{ki}$     &The $i$-th filter in the $k$-th layer  \\\hline

$\bar{\mathbf{w}}_k$     &The exemplar filters in the $k$-th layer \\\hline

$\bar{\mathbf{w}}_{ki}$    &The $i$-th exemplar filter in the $k$-th layer  \\\hline

$s_k(i, j)$     &\tabincell{c}{How well the filter $\mathbf{w}_{kj}$ is suited to be  the exemplar \\ of the filter $\mathbf{w}_{ki}$}       \\\hline

$r_k(i, j)$    &\tabincell{c}{The ``responsibility'' message in the Affinity Propagation}     \\\hline

$a_k(i, j)$         &\tabincell{c}{The ``availability'' message in the Affinity Propagation}     \\\hline
\end{tabular}
\end{table}

For the ease of the following context, we rewrite the parameter tensor of $\mathbf{w}_k$ as a 2-D data matrice with the shape of $c_k \times (c_{k-1} \cdot h_k \cdot w_k)$. And then, we append the biases of the filters to $\mathbf{w}_k$, thus its dimension becomes $c_k \times (c_{k-1} \cdot h_k \cdot w_k + 1)$. Without loss of generality, we denote the $i$-th filter as $\mathbf{w}_{ki}$.

The goal of filter pruning is to obtain a compressed representation of the parameters $\tilde{\mathbf{W}} = \{ \tilde{\mathbf{w}}_k \}_{k=1}^L$ and remove the filters of informational redundancy. Each $\tilde{\mathbf{w}}_k$ has a smaller shape of $\tilde{c}_{k} \times (\tilde{c}_{k-1} \cdot h_k \cdot w_k + 1)$ and $\tilde{c}_{k} \le c_{k}$, $\tilde{c}_{k-1} \le c_{k-1}$, which would be computationally efficient. To this end, predominant works follow: (1) Human designated pruned architecture and inheriting the most important filter weights via rules \cite{he2017channel,lin2020hrank}. (2) Training from scratch with hand-crafted constraints \cite{huang2018data,lin2019towards}. (3) Searching optimal pruned architecture via heuristic-based search algorithm \cite{liu2019metapruning,lin2020channel}. However, these methods suffer either human effort, or heavy computation as discussed in Sec.\,\ref{introduction}.
We propose EPruner towards solving the above problems as stated in Sec.\,\ref{introduction}. The goal of our EPruner is to select $\tilde{c}_k$ high-quality exemplar filters for the $k$-th layer, denoted as $\bar{\mathbf{w}}_k$ from $\mathbf{w}_k$ by the affinity propagation \cite{frey2007clustering}. The exemplars $\bar{\mathbf{w}}_k$ ($k=1,2,...,L$) has the shape of $\tilde{c}_k \times (c_{k-1} \cdot h_k \cdot w_k + 1)$ and $\tilde{c}_k$ makes up the pruned network architecture which will be initialized for the end-to-end fine-tuning. Our method is directly applied on top of the CNN weights, thus the specific value of $\tilde{c}_k$ is adaptive to the input CNNs. Besides, it requires neither human designation, nor search progress in our EPruner. 

We summarize the main notations used in this paper in Tab.\,\ref{notation}, more of which will be detailed in the following contents.

\subsection{The Proposed EPruner}\label{exemplar}
Affinity propagation \cite{frey2007clustering} was originally proposed to select exemplars for the data points with different properties. Early works on the network interpretation \cite{li2015convergent,morcos2018importance,zhou2018revisiting} reveal that the filters, even within the same layer, have different impacts to the network. This indicates that different filters have different properties and there exist some exemplars which contribute more to the network.

As illustrated in Fig.\,\ref{framework}, we propose to regard each filter $\mathbf{w}_{ki}$ as a high-dimensional data point by reformatting it in a vector form, \emph{i.e.}, $\mathbf{w}_{ki} \in \mathbb{R}^{{c}_{k-1} \cdot h_k \cdot w_k}$. For any two filters $\mathbf{w}_{ki}$ and $\mathbf{w}_{kj}$, affinity propagation takes as input their similarity graph $s_k(i, j)$ which reflects how well the filter $\mathbf{w}_{kj}$ is suited to be the exemplar of the filter  $\mathbf{w}_{ki}$. We found our method can well perform with the naive negative Euclidean distance as:
\begin{equation}\label{similarity}
\begin{split}
s_k(i, j)  = - \| \mathbf{w}_{ki} - \mathbf{w}_{kj} \|^2 \quad s.t. \;\;1 \le i, j \le c_k,\;i \ne j.
\end{split}
\end{equation}

When $i = j$, it indicates the suitedness of filter $\mathbf{w}_i$ to be the exemplar of itself (self-similarity). Following \cite{frey2007clustering}, it can be defined as: 
\begin{equation}\label{old_preference}
s_k(i, i)  = \text{median}(\mathbf{w}_{k}),
\end{equation}
where $\text{median}(\cdot)$ returns the median value of the input.

Larger $s_k(i, i)$ leads to more exemplar filters, which however returns less complexity reduction. Using the median value of the whole weight in the $k$-th layer ($\mathbf{w}_k$) in Eq.\,\ref{old_preference} would result in a moderate number of exemplars. To solve, we reformulate Eq.\,\ref{old_preference} as follows:
\begin{equation}\label{new_preference}
\begin{split}
s_k(i, i)  = \beta * \text{median}(\mathbf{w}_{ki}) \qquad s.t. \;\; 0 < \beta \le 1,
\end{split}
\end{equation}
where $\beta$ is a pre-given hyper-parameter.

Eq.\,\ref{new_preference} differs from Eq.\,\ref{old_preference} in two-fold: First, the median value is obtained upon the $i$-th filter $\mathbf{w}_{ki}$ rather than the whole weight $\mathbf{w}_k$. Thus, the similarity $s_k(i, i)$ can be more adaptive to the filter $\mathbf{w}_{ki}$. Second, the introduced $\beta$ provides an adjustable reduction of model complexity. As shown in Sec.\,\ref{ablation}, large $\beta$ leads to a high complexity reduction, and vice versa. 

Besides the similarity, there are also two kinds of messages passing between filters,  \emph{i.e.}, ``responsibility'' and ``availability'', to decide which filters are exemplars, and for every other filter, which exemplar it belongs to.

The ``responsibility'' $r_k(i, j)$ indicates how well the filter $\mathbf{w}_{kj}$ is suited to serve as the exemplar of the filter $\mathbf{w}_{ki}$ by considering other potential exemplars for filter $\mathbf{w}_{ki}$. The updating of $r(i, j)$ follows:
\begin{equation}\label{r}
\begin{split}
r(i, j) \leftarrow  s&(i,  j) \, - \mathop{\max}\limits_{j' s.t. j' \neq j} \big( a(i, j') + 
s(i, j')\big) \\& s.t. \;\;1 \le i, j \le c_k, \; i \ne j,
\end{split}
\end{equation}
where $a(i, j)$ is the ``availability'' below and initialized to zero. 
Initially, $r(i,j)$ is set to $s(i, i)$ minus the largest of the similarities between filter $\mathbf{w}_{ki}$ and other filters. 
Later, if one filter is assigned to other exemplars, its availability is smaller than zero per Eq.\,(\ref{a}) below, which further decreases the effectiveness of $s(i, j')$ in Eq.\,(\ref{r}), thus $\mathbf{w}_{kj'}$ is removed from the exemplar candidates.

For $i = j$, the ``self-responsibility'' is given as:
\begin{equation}\label{rii}
r(i, i) \leftarrow s(i, i) \, - \mathop{\max}\limits_{i' s.t. i' \neq i} s(i, i'),
\end{equation}
which is set to $s(i, i)$ minus the largest of the similarities between filter $\mathbf{w}_{ki}$ and other filters. It reflects possibility that filter $\mathbf{w}_{ki}$ can be an exemplar by considering how ill-suited it is to be assigned to another exemplar.

As for the ``availability'', we first give its updating rule as:
\begin{equation}\label{a}
\begin{split}
a(i, j) \leftarrow \min  \{0, & r(j, j) + 
\mathop{\sum}\limits_{i'\, s.t. \, i' \notin \{ i, j \}} \max \big(0, r(i', j) \big) \}   \\& s.t. \; 1 \le i, j \le c_k, \; i \ne j.
\end{split}
\end{equation}

The availability $a(i,j)$ is set as $r(j, j)$ plus the sum of other responsibilities that filter $\mathbf{w}_{kj}$ receives from others. The $\max()$ excludes the negative responsibilities since we only need to focus on the good filters (positive responsibilities). The $r(j,j) < 0$ denotes that filter $\mathbf{w}_{kj}$ is more appropriate to belong to another exemplar rather than being an exemplar itself. It can be seen that the availability of filter $\mathbf{w}_{kj}$ as an exemplar increases if some other filters have positive responsibilities for $\mathbf{w}_{kj}$. Thus, the ``availability'' $a_k(i, j)$ reflects how appropriate filter $\mathbf{w}_{ki}$ chooses filter $\mathbf{w}_{kj}$ as its exemplar by considering the support from other filters that filter $\mathbf{w}_{kj}$ should be an exemplar. Lastly, the $\min()$ limits the influence of strong positive responsibilities, such that the total sum cannot go above zero.

For $i = j$, the ``self-availability'' is defined as:

\begin{equation}\label{aii}
a(i, i) \leftarrow \mathop{\sum}\limits_{i' \, s.t., i' \, \neq i} \max \big(0, r(i', i)\big),
\end{equation}
which reflects suitedness that filter $\mathbf{w}_{ki}$ is an exemplar, based on the positive responsibilities from other filters.

The updating of ``responsibility'' and ``availability'' is iterative. To avoid the numerical oscillations, we consider the weighted sum for each message at the $t$-th updating stage :
\begin{equation}\label{update_r}
r^t (i, j) = \lambda * r^{(t-1)}(i, j) + (1 - \lambda) * r^t(i,j),
\end{equation}

\begin{equation}\label{update_a}
a^t (i, j) = \lambda * a^{(t-1)}(i, j) + (1 - \lambda) * a^t(i,j),
\end{equation}
where $0 \le \lambda \le 1$ is a weighted factor and is set to 0.5 in our experiments.

After a fixed number of iterations (200 in our experiments), filter $\mathbf{w}_{ki}$ selects as its exemplar another filter $\mathbf{w}_{kj}$ that satisfies:
\begin{equation}\label{final}
\begin{split}
\mathop{\arg\max}\limits_{j} \; r(i, j) + a(i, j)  \qquad s.t. \;\; 1 \le j \le c_k.
\end{split}
\end{equation}

When $i = j$, filter $\mathbf{w}_{ki}$ selects itself as the exemplar. All selected filters make up the exemplars $\bar{\mathbf{w}}_k$. Thus, the number of exemplars, \emph{i.e.}, $\tilde{c}_k$, is self-adaptive without the human designation. In Sec.\,\ref{ablation}, we demonstrate that the exemplar selection can be efficiently implemented on a single CPU, leading to a magnitude-order reduction of time consumption.


\begin{figure*}[!t]
\begin{center}
\includegraphics[height=0.4\linewidth]{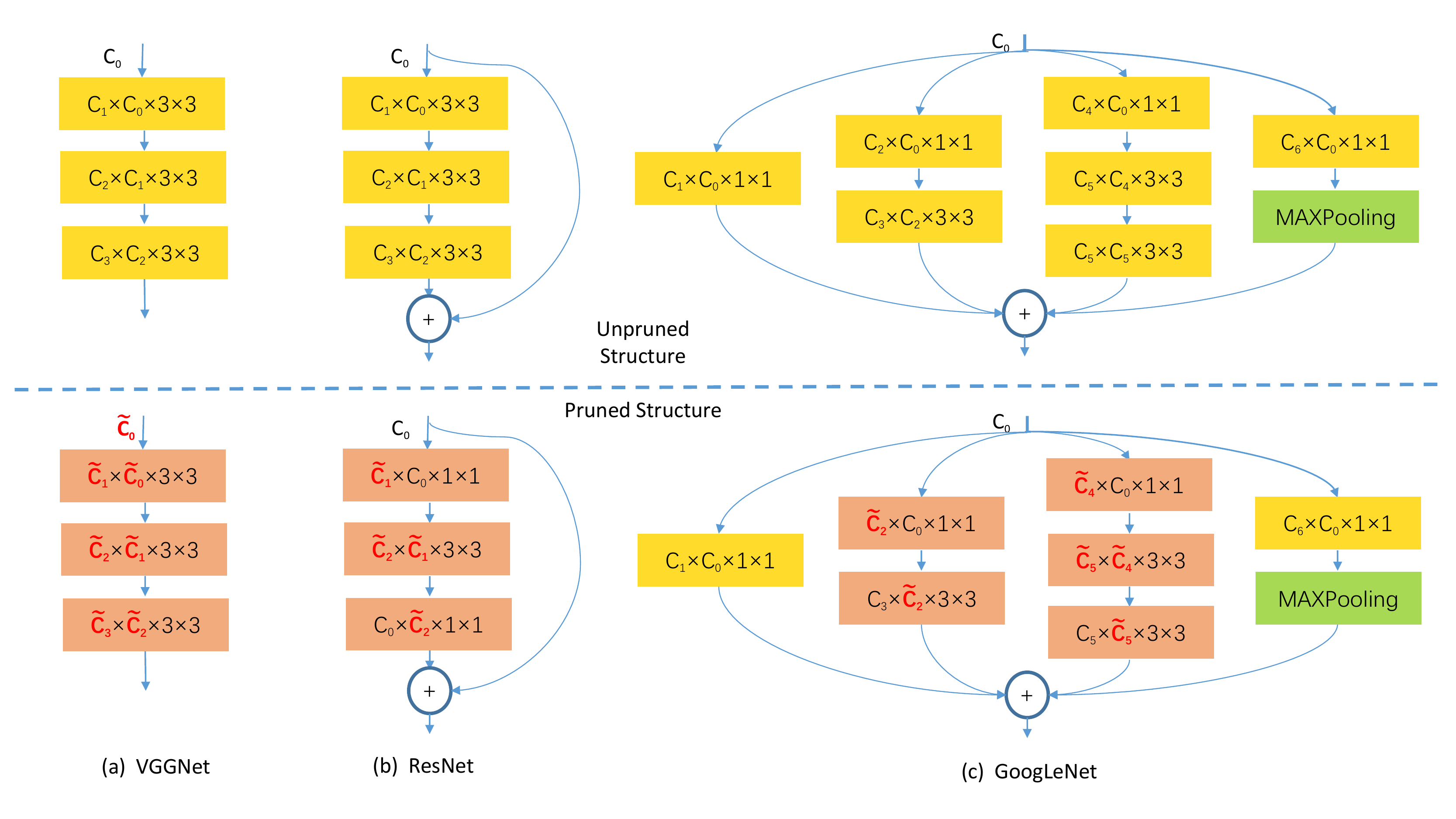}
\end{center}
\caption{\label{pruning}
Pruning strategy for VGGNet, ResNet and GoogLeNet in our EPruner. For VGGNet, all convolutional layers are pruned. For ResNet, we only prune the first two convolutional layers in each residual block. For GoogLeNet, we only prune the branches with more than one convolutional layer. For all pruned filters, the channels in the next layer are removed correspondingly. (Best viewed with zooming in)
}
\end{figure*}

\subsection{Weight Initialization}\label{weight_init}
The pruned network architecture consists of the number of exemplar filters in each layer, \emph{i.e.}, $\tilde{c}_i$. The fine-tuning is required to recover the accuracy of pruned network such that it would keep a better or at least comparable performance against the pre-trained model. Thus, the key now falls in how to feed a good weight initialization to the pruned network architecture for the follow-up fine-tuning. To that effect, we consider four different weight initialization scenarios:

(1) Exemplar Weights. The weights of exemplar filters are regarded as the initial weights. To ensure the dimension consistency between the exemplar filters $\bar{\mathbf{w}}_k$ with the shape of $\tilde{c}_k \times (c_{k-1} \cdot h_k \cdot w_k + 1)$ and pruned network parameters $\tilde{\mathbf{w}}_k$ with the shape of $\tilde{c}_k \times (\tilde{c}_{k-1} \cdot h_k \cdot w_k + 1)$, the corresponding channels in the weights of exemplars are removed directly.

(2) Random Projection. Instead of directly removing the corresponding channels in the weights of exemplars, we apply the sparse random projection \cite{li2006very} to reduce the dimension of exemplars, which is able to preserve the similarity between data points as demonstrated in \cite{sulic2010efficient}.

(3) $\ell_1$-norm weights. Filters with large norms are believed to have the ability to retain more information, thus the corresponding weights are inherited as a warm-up for the follow-up fine-tuning~\cite{li2017pruning}. Such a scenario has been widely-known in the network pruning.

(4) Random Initialization. Recent progress \cite{liu2019rethinking} shows that the essence of filter pruning lies in finding the optimal pruned architecture and random initialization with Gaussian distribution can perform better than inheriting the most important filters. Thus, we also consider random Gaussian distribution.

We use the exemplar weights by default. In Sec.\,\ref{initialization}, we conduct ablation study regarding these different ways of initialization and justify the correctness of inheriting the most important filters in the literature.

\section{Experiments}\label{experiment}
We apply the proposed EPruner on the CIFAR-10 benchmark \cite{krizhevsky2009learning} using three classic deep networks including VGGNet \cite{simonyan2015very}, GoogLeNet \cite{szegedy2015going} and ResNets \cite{he2016deep}, and the ILSVRC-2012 benchmark \cite{russakovsky2015imagenet} using ResNets \cite{he2016deep} with different depths. VGGNet contains sequential convolutions, GoogLeNet has the convolutions with multiple branches, and ResNet is especially designed with residual blocks. Fig.\,\ref{pruning} outlines our pruning strategy for the three kinds of networks. We carry out our experiments on NVIDIA Tesla V100 GPUs. All models are implemented and trained using Pytorch \cite{paszke2017automatic}.

\begin{table*}[!t]
\centering
\caption{\label{cifar}We apply the proposed EPruner to VGGNet-16 \protect\cite{simonyan2015very}, GoogLeNet \protect\cite{szegedy2015going} and ResNets with different depths of 56 and 110 \protect\cite{he2016deep} and evaluate these models on the CIFAR-10 benchmark. The channels, FLOPs, parameters and Top-1 accuracy are reported. The numerical value after EPruner denotes the value of $\beta$.}
\begin{tabular}{ccccccccc}
\hline
Model    &Top1-acc &$\uparrow\downarrow$ &Channels  &Pruning Rate &FLOPs  &Pruning Rate  &Parameters &Pruning Rate\\ \hline
VGGNet-16     &93.02\%    &0.00\%    &4224  &0.00\%    &314.59M  &0.00\% &14.73M     &0.00\%    \\
VGGNet-16 0.5$\times$ &92.68\%&0.34\%$\downarrow$&2112&50.00\%&79.20M&74.97\%&3.69M&74.79\% \\
\textbf{EPruner-0.73} &\textbf{93.08\%}&\textbf{0.06\%$\uparrow$}&\textbf{1363}&\textbf{67.73\%}&\textbf{74.42M}&\textbf{76.34\%}&\textbf{1.65M}&\textbf{88.80\%} \\ \hline
GoogLeNet&95.05\%  &0.00\%  &7904  &0.00\% &1534.55M &0.00\% &6.17M   &0.00\%    \\
GoogLeNet 0.25$\times$ &94.38\%&0.67\%$\downarrow$&6236&21.10\%&587.20M&61.61\%&2.61M&57.72\%\\
\textbf{EPruner-0.65} &\textbf{94.99\%}&\textbf{0.06\%$\downarrow$} &\textbf{6110}&\textbf{22.70\%}&\textbf{500.87M}&\textbf{67.36\%}  &\textbf{2.22M}&\textbf{64.20\%} \\ \hline
ResNet-56  &93.26\%    &0.00\%   &2032 &0.00\%  &127.62M  &0.00\% &0.85M    &0.00\%    \\
ResNet-56 0.5$\times$ &91.90\% &1.36\%$\downarrow$ &1528 &24.80\% &63.80M &49.61\% &0.43M &49.82\% \\
\textbf{EPruner-0.76}&\textbf{93.18\%} &\textbf{0.08\%$\downarrow$}  &\textbf{1450}&\textbf{28.64\%} &\textbf{49.35M} &\textbf{61.33\%} &\textbf{0.39M}  &\textbf{54.20\%} \\ \hline
ResNet-110 &93.50\%    &0.00\%     &4048   &0.00\% &257.09M  &0.00\% &1.73M   &0.00\%    \\
ResNet-110 0.4$\times$ &92.69\%&0.81\%$\downarrow$&2806&30.68\%&97.90M&61.13\%&0.67M&61.62\% \\
\textbf{EPruner-0.60} &\textbf{93.62\%}&\textbf{0.12\%$\uparrow$} &\textbf{2580} &\textbf{36.26\%}  &\textbf{87.65M}  &\textbf{65.91\%}  &\textbf{0.41M}  &\textbf{76.30\%}\\ \hline
\end{tabular}
\end{table*}

\subsection{Implementation Details}\label{implementation}
\textbf{Training Settings}.
We fine-tune all models by using Stochastic Gradient Descent (SGD) optimizer with a momentum of 0.9 and the batch size is set to 256. Also, on CIFAR-10, we use a weight decay of 5$\times$10$^{\text{-3}}$ and 150 epochs are given for fine-tuning. The learning rate starts from 0.01 and is reduced by a factor of 10 after 50 and 100 epochs. On ILSVRC-2012, the weight decay  is set to 1$\times$10$^\text{-4}$ and we fine-tune the networks for 90 epochs. The learning rate is initially set to 0.1 and divided by 10 every 30 epochs. Without specifications, for all methods, we apply the random crop and horizontal flip to the input images, which are also official operations in Pytorch. To stress, other techniques for image augmentation, such as lightening and color jitter, can be applied to further improve the accuracy performance as done in the source codes of \cite{yu2019slimmable,liu2019metapruning,ding2019approximated}. Besides, the cosine scheduler for learning rate can also be used to replace the step learning, which also has been demonstrated to be able to boost the performance in \cite{luo2020autopruner,he2018amc,li2020eagleeye}. We do not consider these in this paper since we aim to show the performance of pruning algorithms themselves.
\textbf{Performance Metrics}.
We report channels, FLOPs (floating-point operations) and parameters to measure the effect of pruning. Channels reflect the memory footprint. FLOPs and parameters reflect the computation cost and storage space, respectively. Also, their corresponding pruning rates are reported. Besides, for CIFAR-10, top-1 accuracy of pruned models is provided. For ILSVRC-2012, both top-1 and top-5 accuracies are reported.

\subsection{Testing Results on CIFAR-10}\label{results_cifar}
Tab.\,\ref{cifar} displays our pruning results for VGGNet, GoogLeNet and ResNets on CIFAR-10. More detailed analyses are provided as below:

\textbf{VGGNet}.
We choose to prune the 16-layer VGGNet model on CIFAR-10. As can be seen from Tab.\,\ref{cifar}, EPruner greatly simplifies the model complexity by reducing 67.73\% channels, 76.34\% FLOPs and 88.80\% parameters, meanwhile it still obtains an accuracy improvement of 0.06\% (93.08\% for EPruner \emph{vs.} 93.02\% for the baseline). This greatly facilitates VGGNet model, a popular backbone for object detection and semantic segmentation, to be deployed on mobile devices. Hence, EPruner demonstrates its ability to compress and accelerate the neural network with the sequential structure.

\textbf{GoogLeNet}.
For GoogLeNet, as shown in Tab.\,\ref{cifar}, EPruner removes 22.70\% channels, 67.36\% FLOPs and 64.20\% parameters with a negligible accuracy drop (94.99\% for EPruner \emph{vs.} 95.05\% for the baseline). Besides, we observe that for GoogLeNet, the channel reduction is not so high as that in VGGNet. To analyze, as shown in Fig.\,\ref{pruning}, we do not prune the branch with only one convolutional layer and the last convolutional layer in the branch with more than one layer is also not pruned, which explains the above observation. Nevertheless, as can be seen, the reductions of FLOPs and parameters are still significant. Hence, EPruner can be well applied in the neural network with a multi-branch structure.

\textbf{ResNets}.
To evaluate the power of EPruner in complexity reduction for network with residual blocks. We choose to prune ResNets with different depths of 56 and 110. As shown in Tab.\,\ref{cifar}, EPruner boosts the computation by decreasing about 61.33\% FLOPs for ResNet-56 and 65.91\% FLOPs for ResNet-110, and it also saves more than half storage space by reducing 54.20\% parameters for ResNet-56 and 76.30\% parameters for ResNet-110. Comparing ResNet-56 with ResNet-110, we observe that more FLOPs and parameters are reduced in ResNet-110. The potential reason might be that the deeper ResNet-110 suffers more over-parameterized burden. Besides, it can also be observed that, similar to GoogLeNet, the channel reduction for ResNets is also limited. As explained in Fig.\,\ref{pruning}, we do not prune the convolution in the skip connection and the last convolutional layer in the residual block. Lastly, we can see that EPruner can well maintain the accuracy performance of the baseline model. For ResNet-56, EPruner leads to little accuracy loss (93.18\% for EPruner \emph{vs}. 93.26\% for the baseline), while for ResNet-110, EPruner gains an obvious accuracy improvement of 0.12\%. The above observations verify the effectiveness of EPruner in compressing and accelerating the residual-designed networks.

Besides, Tab.\,\ref{cifar} also displays the results of uniform pruning. The digit after the network denotes the percentage of preserved filters in each layer of the network. As can be seen, our EPruner outperforms uniform pruning regarding the accuracy performance even with more reductions of the model complexity, which well demonstrates that EPruner can return an adaptive pruned network architecture with a better performance.

Also, we observe that the performance of pruned VGGNet-16 and ResNet-110 is better than that of the models before pruning. The rationale behind this is that pruning a big neural network to a smaller one also has the advantage of avoiding overfitting and improving generalization~\cite{sum1999kalman,sum1999adaptive}. The VGGNet-16 and ResNet-110 are two over-parameterized networks, which often result in an over-fitting problem when trained on the small CIFAR-10. Our network pruning helps to relieve overfitting and improve generalization as discussed above, and thus better performance can be observed.

\begin{table*}[!t]
\scriptsize
\centering
\caption{\label{imagenet}We apply the proposed EPruner to ResNets with different depths of 18, 34, 50, 101 and 152 \protect\cite{he2016deep} and evaluate these models on the ILSVRC-2012 benchmark. The channels, FLOPs, parameters, Top-1 and Top-5 accuracies are reported. The numerical value after EPruner denotes the value of $\beta$.}
\begin{tabular}{ccccccccccc}
\hline
Model   &Top1-acc &$\uparrow\downarrow$ &Top5-acc &$\uparrow\downarrow$ &Channels &Pruning Rate  &FLOPs  &Pruning Rate  &Parameters &Pruning Rate \\ \hline
ResNet-18 &69.66\%&0.00\%&89.08\%&0.00\%&4800&0.00\% &1824.52M&0.00\% &11.69M&0.00\% \\
ResNet-18 0.55$\times$ &66.93\%&2.73\%$\downarrow$&87.21\%&1.87\%$\downarrow$&3932&18.08\%&1059.59M&41.75\%&6.73M&42.46\% \\
\textbf{EPruner-0.73}&\textbf{67.31\%}&\textbf{2.35\%$\downarrow$}&\textbf{87.70\%}&\textbf{1.38\%$\downarrow$}&\textbf{3888}&\textbf{19.00\%}&\textbf{1024.01M}&\textbf{43.88\%}&\textbf{6.05M}&\textbf{48.25\%} \\ \hline
ResNet-34 &73.28\%&0.00\%&91.45\%&0.00\%&8512&0.00\%&3679.23M&0.00\%&21.90M&0.00\% \\
ResNet-34 0.5$\times$ &69.98\%&3.30\%$\downarrow$&88.89\%&2.56\%$\downarrow$&6624&22.18\%&1906.87M&48.06\%&11.25M&48.39\% \\
\textbf{EPruner-0.75}&\textbf{70.95\%}&\textbf{2.23\%$\downarrow$}&\textbf{89.97\%}&\textbf{1.48\%$\downarrow$}&\textbf{6684}&\textbf{21.48\%}&\textbf{1853.92M}&\textbf{49.61\%}&\textbf{10.24M}&\textbf{53.24\%}\\ \hline
ResNet-50           &76.01\%&0.00\%&92.96\%&0.00\%&26560&0.00\%&4135.70M&0.00\%&25.56M&0.00\% \\
ResNet-50 0.55$\times$ &74.00\%&2.01\%$\downarrow$&91.43\%&1.53\%$\downarrow$&23144&12.86\%&2015.99M&50.97\%&13.40M&47.56\% \\
\textbf{EPruner-0.73}&\textbf{74.26\%}&\textbf{1.75\%$\downarrow$}&\textbf{91.88\%}&\textbf{1.08\%$\downarrow$}&\textbf{22955}&\textbf{13.57\%}&\textbf{1929.15M}&\textbf{53.35\%}&\textbf{12.70M}&\textbf{50.31\%}  \\ \hline
ResNet-101  &77.38\%&0.00\%&93.59\%&0.00\%&52672&0.00\%&7868.40M&0.00\%&44.55M&0.00\% \\
ResNet-101 0.45$\times$ &74.30\%&3.08\%$\downarrow$&92.01\%&1.58\%$\downarrow$&43710&17.01\%&2841.60M&63.73\%&17.42M&60.89\% \\
\textbf{EPruner-0.67}&\textbf{75.45\%}&\textbf{1.93\%$\downarrow$}&\textbf{92.70\%}&\textbf{0.89\%$\downarrow$}&\textbf{42843}&\textbf{18.66\%}&\textbf{2817.27M}&\textbf{64.20\%}&\textbf{15.55M}&\textbf{65.10\%} \\ \hline
ResNet-152          &78.31\%&0.00\%&93.99\%&0.00\%&75712&0.00\%&11605.91M&0.00\%&60.19M&0.00\% \\
ResNet-152 0.45$\times$ &75.91\%&2.40\%$\downarrow$&92.83\%&1.16\%$\downarrow$&62516&17.43\%&4030.03M&65.13\%&22.42M&62.75\% \\
\textbf{EPruner-0.63}&\textbf{76.83\%}&\textbf{1.48\%$\downarrow$}&\textbf{93.28\%}&\textbf{0.71\%$\downarrow$}&\textbf{61688}&\textbf{18.52\%}&\textbf{4047.69M}&\textbf{65.12\%}&\textbf{21.56M}&\textbf{64.18\%}\\ \hline
\end{tabular}
\end{table*}

\subsection{Testing Results on ILSVRC-2012}\label{results_ilsvrc}
Tab.\,\ref{imagenet} presents the pruning results of the proposed EPruner using ResNets with different depths of 18/34/50/101/152 on the large-scale ILSVRC-2012.

As can be observed from Tab.\,\ref{imagenet}, the accuracy performance for compressed models on ILSVRC-2012 generates more compared with that on CIFAR-10. For an in-depth analysis, ILSVRC-2012 is a large-scale benchmark with 1,000 categories which poses a greater challenge, compared with small-scale CIFAR-10 containing only 10 categories, to recognize the images using the compressed models. Nevertheless, as can be seen, the drops of accuracies are still tolerable. Besides, we observe that the efficacy of EPruner is more advantageous in compressing deeper networks. For example, in the deepest ResNet-152, EPruner achieves a 65.12\%-FLOPs reduction by removing 64.18\% parameters and 18.52\% channels, with only 1.48\% top-1 and 0.71\% top-5 accuracy loss. While in ResNet-18, 43.88\% FLOPs 48.25\% parameters and 10.00\% channels are removed with more accuracy drops of 2.35\% in the top-1 and 1.38\% in the top-5. To explain, the shallow ResNet-18 has less redundancies thus more accuracy drops occur even with less reduction of model complexity. Nevertheless, this observation is meaningful since deeper networks are more constrained in the resource-limited environment.

Moreover, similar to the observation in Tab.\,\ref{cifar}, EPruner also advantages in its supreme performance in comparison with the uniform pruning even on the large-scale ILSVRC-2012, which again well demonstrates the effectiveness of our EPruner in reducing the model complexity while retains a better performance.

\begin{table}[!t]
\centering
\caption{\label{other}We compare the proposed EPruner with several SOTAs using ResNet-50 \protect\cite{he2016deep} on ILSVRC-2012, including ThiNet \protect\cite{luo2017thinet}, CP \protect\cite{he2017channel}, SSS \protect\cite{huang2018data}, GAL \protect\cite{lin2019towards}, MetaPruning \protect\cite{liu2019metapruning}, HRank \protect\cite{lin2020hrank} and ABCPruner \protect\cite{lin2020channel}. Following \cite{liu2019metapruning,lin2020channel}, the FLOPs, top-1 accuracy and the training/fine-tuning epochs of the compressed model are reported.}
\begin{tabular}{c|c|c|c}
\hline
Model           &FLOPs &Top1-acc          &Top5-acc\\ \hline
ThiNet-30 \cite{luo2017thinet}  & 1.10B  & 68.42\%  &88.30\%\\
MetaPruning-0.50 \cite{liu2019metapruning} &1.03B  &69.92\%    &89.60\%\\
HRank \cite{lin2020hrank} &0.98B &69.10\%  &89.58\%\\
ABCPruner-30\% \cite{lin2020channel} &0.94B &70.29\%  &89.63\%\\
\textbf{EPruner-0.85} &\textbf{0.91B} &\textbf{70.34}\% &\textbf{89.65\%} \\
SSS-26 \cite{huang2018data} &2.33B   &71.82\%   &90.79\%\\
GAL-0.5 \cite{lin2019towards}    & 2.33B  & 71.95\%   &90.94\% \\
GAL-0.5-joint \cite{lin2019towards} & 1.84B  &71.80\%   &90.82\%\\
ThiNet-50 \cite{luo2017thinet}  & 1.71B  & 71.01\%   &90.02\%\\
MetaPruning-0.75 \cite{liu2019metapruning} & 2.26B  & 72.17\%    &90.86\% \\
HRank \cite{lin2020hrank} &1.55B &71.98\%  &91.01\%\\
ABCPruner-50\% \cite{lin2020channel} &1.30B &72.58\%&90.91\%\\
\textbf{EPruner-0.81} &\textbf{1.29B} &\textbf{72.73\%} &\textbf{91.01\%} \\
SSS-32 \cite{huang2018data}  &2.82B   &74.18\%  &91.91\%\\
CP \cite{he2017channel}   & 2.73B  & 72.30\%    &80.80\%\\
MetaPruning-0.85 \cite{liu2019metapruning} & 2.92B  & 74.49\%    &92.14\% \\
ABCPruner-100\% \cite{lin2020channel} &2.56B &74.84\%   &92.27\%\\ 
\textbf{EPruner-0.71} &\textbf{2.37B} &\textbf{74.95\%}  &\textbf{92.36\%} \\ \hline
\end{tabular}
\end{table}

\subsection{Comparison with Other Methods}\label{comparison}
Further, we compare our EPruner with several SOTAs including rule-based pruning \cite{luo2017thinet,he2017channel,lin2020hrank}, regularization-driven pruning \cite{huang2018data,lin2019towards}, and architecture-search-driven pruning \cite{liu2019metapruning,lin2020channel}. Following \cite{liu2019metapruning,lin2020channel}, the experiments are implemented using ResNet-50 on ILSVRC-2012 and we report the FLOPs, top-1 accuracy, and the training/fine-tuning epochs of the compressed model for all methods in Tab.\,\ref{other}.

From Tab.\,\ref{other}, EPruner takes the lead in both the model accuracy retaining and the FLOPs reduction. When $\beta = 0.85$, EPruner reduces the FLOPs of pre-trained model to 0.91G with the top-1 accuracy of 70.34\% while the architecture-search-driven ABCPruner-30\% remains 0.94G FLOPs with a lower top-1 accuracy of 70.29\%. When $\beta = 0.81$, EPruner shows an overwhelming superiority over the rule-based HRank and regularization-driven GAL (1.29G \emph{vs.} 1.55G \emph{vs.} 1.84G in FLOPs, and 72.73\% \emph{vs}. 71.98\% \emph{vs}. 71.80\% in top-1 accuracy). Compared with ABCPruner-50\% which has the FLOPs of 1.30G and the top-1 accuracy of 72.73\%, the advantages of EPruner are obvious. When $\beta = 0.71$, EPruner has a substantially better performance. Compared to ABCPruner-100\% that has 2.56G FLOPs and accuracy of 74.84\%, EPruner has only 2.37G FLOPs while yielding 74.95\% in top-1 accuracy. Overall, EPruner demonstrates its ability to accelerate the CNNs while maintains a good accuracy performance.

\subsection{In-depth Analysis}\label{indepth}
The experiments in Tab.\,\ref{cifar}, Tab.\,\ref{imagenet} and Tab.\,\ref{other} show that EPruner can be well applied to reduce the complexity of CNNs while keeping a better or at least comparable accuracy performance against the pre-trained models or state-of-the-art methods. The efficacy of our EPruner is mainly from two points: self-adapting to the optimal pruned architecture and inheriting the most important filter weights.

\begin{figure}[!t]
\begin{center}
\includegraphics[height=0.4\linewidth]{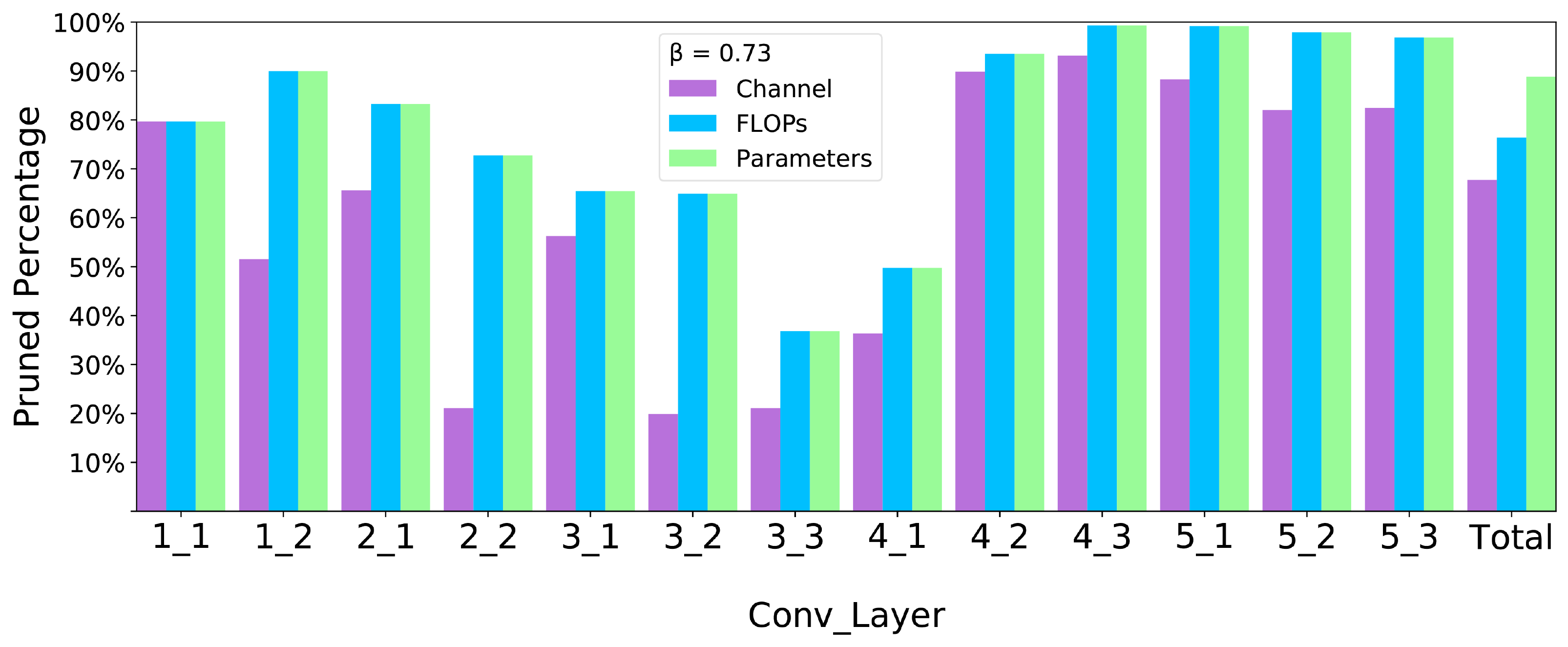}
\end{center}
\caption{\label{ratio}
The pruning rate of each layer for VGGNet on CIFAR-10 when $\beta = 0.73$.
}
\end{figure}

\textbf{Adaptive Pruned Architecture}. Rule-based pruning and regularization-driven require the involvement of human labor to designate the pruned architecture or hyper-parameter analysis, results of which are usually sub-optimal, while architecture-search-driven pruning has to be re-run for some times to pick up the best one. To analyze EPruner, we show the layer-wise pruning when $\beta = 0.73$ using VGGNet-16 in Fig.\,\ref{ratio}. As can be seen, the pruning rate varies across different layers. More filters are preserved in the middle layers (3\_2 to 4\_1) while the other layers tend to remove more filters. By selecting the exemplars via message passing among filters, EPruner self-adapts to the filter property and derives the deterministic optimal pruned architecture without human involvement.

\begin{table}[!t]
\centering
\caption{\label{initialization}Comparisons of pruned architecture with different weight initialization. We set the value of $\beta$ as that in Tab.\,\ref{cifar} and Tab.\,\ref{imagenet} for each network and report the top-1(top-5) accuracy.}
\setlength{\tabcolsep}{0.2em}
\begin{tabular}{c|c|c|c|c}
\hline
             &\tabincell{c}{Exemplar \\ Weights (\%)}    &\tabincell{c}{Random \\ Projection (\%)} &\tabincell{c}{$\ell_1$-norm \\ weights (\%)}  &\tabincell{c}{Random \\ Initialization (\%)} \\ \hline
VGGNet-16    &\textbf{93.08}     &92.95          &92.98      &92.61     \\ \hline
GoogLeNet    &\textbf{94.99}     &94.49          &94.41 &94.19     \\ \hline
ResNet-56    &\textbf{93.18}     &92.44          &93.03         &91.45      \\ \hline
ResNet-110   &\textbf{93.62}     &93.02          &92.99           &92.44     \\ \hline
ResNet-18    &\textbf{67.31}(\textbf{87.42}) &66.68(87.45) &67.01(87.42)   &66.46(87.13)   \\ \hline
ResNet-34    &\textbf{70.95}(\textbf{89.97}) &70.79(89.91) &70.76(89.93)  &70.71(89.78)   \\ \hline
ResNet-50    &\textbf{74.26}(\textbf{91.88}) &73.80(91.83) &73.99(91.82) &73.54(91.55)  \\ \hline
ResNet-101   &\textbf{75.45}(\textbf{92.70}) &75.31(92.51) &75.40(92.58) &75.12(92.25)  \\ \hline
ResNet-152  &\textbf{76.51}(\textbf{93.22}) &76.43(93.14)  &76.46(93.20) &76.15(92.97)      \\ \hline
\end{tabular}
\end{table}

\textbf{Important Filter Weights}. Recent work \cite{liu2019rethinking} shows that the essence of filter pruning lies in finding the optimal pruned architecture rather than selecting the most important filter weights as in \cite{hu2016network,li2017pruning,lin2020hrank}. In Tab.\,\ref{initialization}, we show a different observation by feeding the pruned architectures with different initial weights as stated in Sec.\,\ref{weight_init}. From Tab.\,\ref{initialization}, using exemplar weights obtains the best accuracy performance while random projection of exemplar weights and weights with larger $\ell_1$-norm feed back the second best, three of which show consistently higher performance than the random initialization. 
It indicates that the pre-trained exemplar weights are already a ``distilled'' piece of information from the large-scale training data of CNNs, thus they can provide a better warm-up for fine-tuning, which re-justifies the correctness of inheriting the most important filter weights. 
We assume that the criteria for measuring the most important filter weights in previous rule-based pruning methods are not robust and supportive, thus random initialization shows better results \cite{liu2019rethinking}.

\subsection{Ablation Study}\label{ablation}
In this section, we show the influence of $\beta$ and the comparison of practical time consumption in finding out the optimal pruned architecture.

\begin{figure}[!t]
\begin{center}
\begin{minipage}[t]{0.48\linewidth}
\centerline{
\subfigure[$\beta$ ranges from 0 to 1.]{
\includegraphics[width=\linewidth]{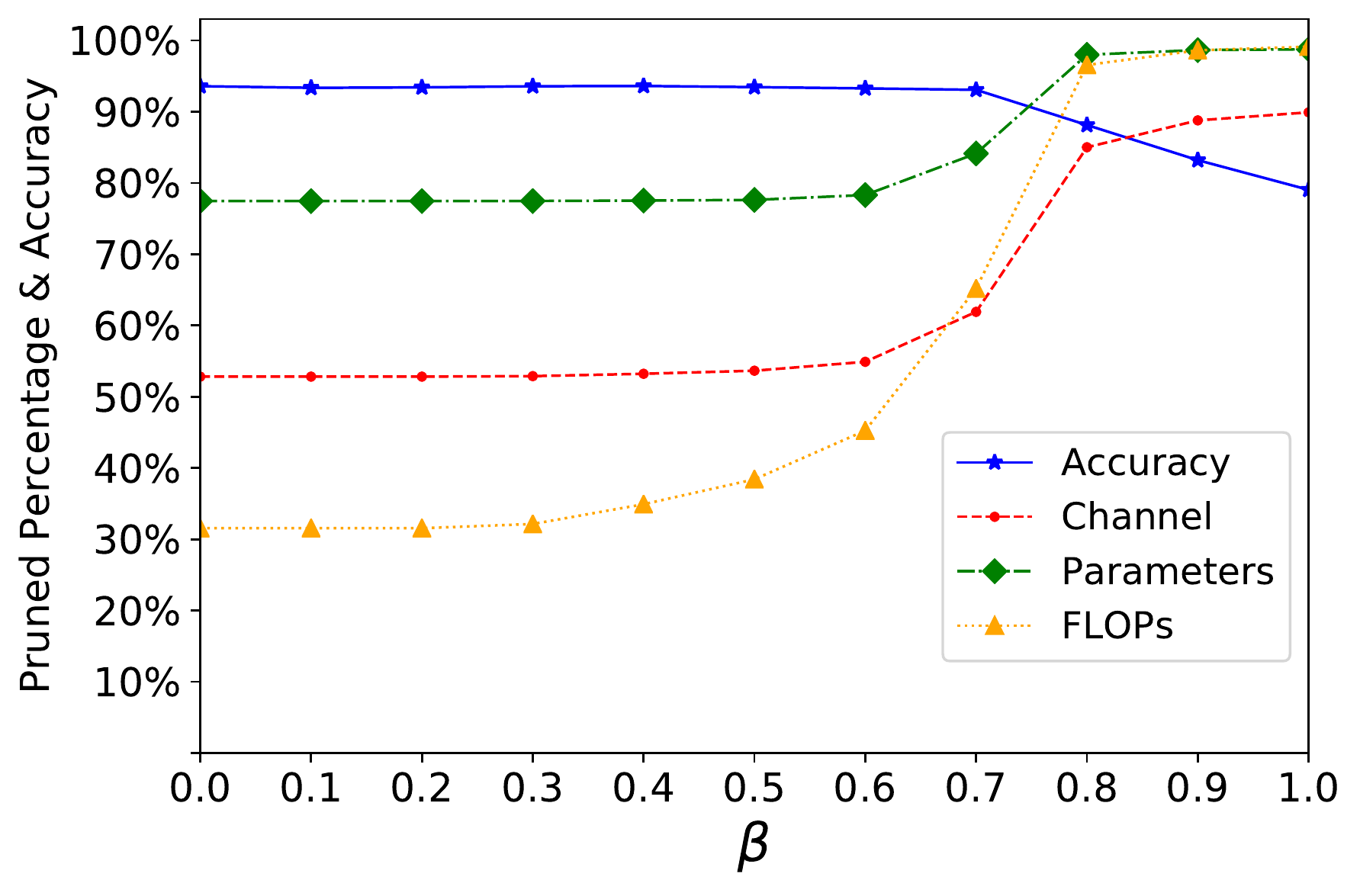}}
\subfigure[$\beta$ ranges from 0.7 to 0.8.]{
\includegraphics[width=\linewidth]{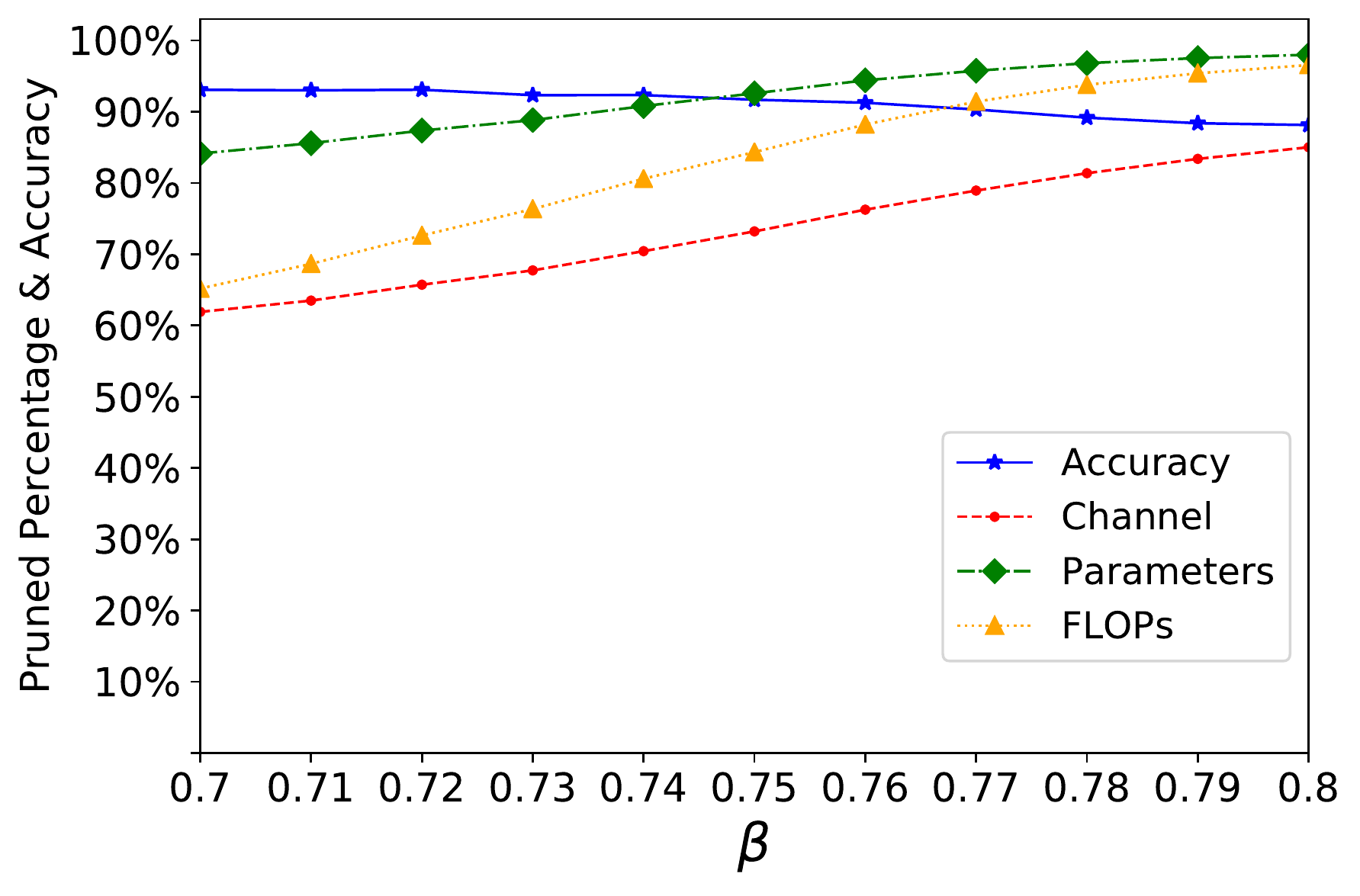}}
}
\end{minipage}

\end{center}
\caption{\label{beta}
The influence of $\beta$ for VGGNet-16 on CIFAR-10. Generally, large $\beta$ obtains higher pruning rates but more accuracy drops.
}
\end{figure}

\textbf{Influence of $\beta$}.
The $\beta$ is used to make the reduction of model complexity more adjustable. As stressed in Sec.\,\ref{exemplar}, smaller $\beta$ leads to more exemplars, thus fewer reductions of model complexity, but better accuracy performance. To demonstrate this, Fig.\,\ref{beta} shows the influence of $\beta$ by using VGGNet-16 on CIFAR-10. Fig.\,\ref{beta}(a) verifies our analysis. The accuracy performance is relatively stable when $\beta$ falls within 0.0 to 0.7. It starts to drop as $\beta$ increases, which makes sense since the pruning rate drastically goes up at the point of $\beta$ = 0.7. Then we show a more subtle interval with $\beta$ ranging from 0.7 to 0.8 in Fig.\,\ref{beta}(b). As can be seen, the suitable value of $\beta$ would be between 0.72 and 0.73. Note that, $\beta$ is used to reach the expected reduction of model complexity, which however does not involve in the process of finding the optimal architecture, which lessens human labors.

\begin{table}[!t]
\centering
\caption{\label{efficiency}Comparisons of time consumption on finding out the pruned network architecture between EPruner tested on NVIDIA Tesla V100 GPUs, and EPruner tested on Intel(R) Xeon(R) CPU E5-2620 v4 @2.10GHz.}
\begin{tabular}{c|c|c|c|c}
\hline
             &ABCPruner    &  GPUs  &\textbf{EPruner (Ours)} & CPUs\\ \hline
VGGNet-16    &5387.24s      &    1   &2.92s                    &1  \\ \hline
GoogLeNet    &26967.65s     &    1   &1.16s                     &1  \\ \hline
ResNet-56    &5810.51s      &    1   &0.38s                    &1 \\ \hline
ResNet-110    &10565.27s    &    1   &0.79s                 &1 \\ \hline
ResNet-18   &28537.18s      &    1   &1.22s            &1 \\ \hline
ResNet-34   &39266.24s      &1      &2.27s   &1 \\ \hline
ResNet-50    &43534.72s &2  &2.88s&1  \\ \hline
ResNet-101   &70181.60s &2  &8.53s&1  \\ \hline
ResNet-152    &75272.32s &3  &12.40s&1  \\ \hline
\end{tabular}
\end{table}

\textbf{Practical Time Consumption}. We choose ABCPruner, which shows the best efficiency and effectiveness among compared methods, for comparison. Tab.\,\ref{efficiency} shows the practical time consumption on finding out the pruned architecture between EPruner and ABCPruner. As can be seen, ABCPruner requires a large search time on NVIDIA Tesla V100 GPUs, even multiple GPUs are necessary in deeper networks, \emph{e.g.}, ResNet-152. The consumption would be significantly heavy in other platforms, such as GTX-1080TI GPUs and CPUs, which is unacceptable. In comparison, EPruner can find out the pruned architecture in seconds simply on CPUs, a magnitude-order reduction of time consumption.  To analyze, ABCPruner is a data-driven approach where the training data has to be used to train each potential architecture for some epochs and evaluate its fitness by observing the accuracy, while EPruner selects the exemplars among filters, which is data-independent and thus more efficient.

\section{Conclusions}\label{conclusion}
In this paper, we present a novel filter pruning method, called EPruner, which aims to select exemplars among filters. To that effect, we regard the filters as a set of high-dimensional data points and the affinity propagation is applied to generate high-quality exemplars. The lack of human involvement and parameter analysis makes the implementation simpler and more effective. The optimal architecture with EPruner can be efficiently implemented within a few seconds simply on the CPUs, leading to a magnitude-order reduction of time consumption. We show that the weights of exemplars can serve as a better warm-up for fine-tuning the network, which justifies the correctness of inheriting the most important filter weights. We demonstrate the applicability of EPruner in compressing various networks and its superiorities over competing SOTAs.

%
%
\bibliographystyle{IEEEtran}
\bibliography{egbib}

\begin{thebibliography}{10}
\providecommand{\url}[1]{#1}
\csname url@samestyle\endcsname
\providecommand{\newblock}{\relax}
\providecommand{\bibinfo}[2]{#2}
\providecommand{\BIBentrySTDinterwordspacing}{\spaceskip=0pt\relax}
\providecommand{\BIBentryALTinterwordstretchfactor}{4}
\providecommand{\BIBentryALTinterwordspacing}{\spaceskip=\fontdimen2\font plus
\BIBentryALTinterwordstretchfactor\fontdimen3\font minus
  \fontdimen4\font\relax}
\providecommand{\BIBforeignlanguage}[2]{{%
\expandafter\ifx\csname l@#1\endcsname\relax
\typeout{** WARNING: IEEEtran.bst: No hyphenation pattern has been}%
\typeout{** loaded for the language `#1'. Using the pattern for}%
\typeout{** the default language instead.}%
\else
\language=\csname l@#1\endcsname
\fi
#2}}
\providecommand{\BIBdecl}{\relax}
\BIBdecl

\bibitem{simonyan2015very}
K.~Simonyan and A.~Zisserman, ``Very deep convolutional networks for
  large-scale image recognition,'' in \emph{Proceedings of the International
  Conference on Learning Representations (ICLR)}, 2015.

\bibitem{chen2017deep}
X.~Chen, J.~Weng, W.~Lu, J.~Xu, and J.~Weng, ``Deep manifold learning combined
  with convolutional neural networks for action recognition,'' \emph{IEEE
  Transactions on Neural Networks and Learning Systems (TNNLS)}, vol.~29,
  no.~9, pp. 3938--3952, 2017.

\bibitem{long2015fully}
J.~Long, E.~Shelhamer, and T.~Darrell, ``Fully convolutional networks for
  semantic segmentation,'' in \emph{Proceedings of the IEEE Conference on
  Computer Vision and Pattern Recognition (CVPR)}, 2015, pp. 3431--3440.

\bibitem{peng2019few}
Z.~Peng, Z.~Li, J.~Zhang, Y.~Li, G.-J. Qi, and J.~Tang, ``Few-shot image
  recognition with knowledge transfer,'' in \emph{Proceedings of the IEEE
  Conference on Computer Vision and Pattern Recognition (CVPR)}, 2019, pp.
  441--449.

\bibitem{denton2014exploiting}
E.~L. Denton, W.~Zaremba, J.~Bruna, Y.~LeCun, and R.~Fergus, ``Exploiting
  linear structure within convolutional networks for efficient evaluation,'' in
  \emph{Proceedings of the Advances in Neural Information Processing Systems
  (NeurIPS)}, 2014, pp. 1269--1277.

\bibitem{hayashi2019exploring}
K.~Hayashi, T.~Yamaguchi, Y.~Sugawara, and S.-i. Maeda, ``Exploring unexplored
  tensor network decompositions for convolutional neural networks,'' in
  \emph{Proceedings of the Advances in Neural Information Processing Systems
  (NeurIPS)}, 2019, pp. 5552--5562.

\bibitem{kim2019efficient}
H.~Kim, M.~U.~K. Khan, and C.-M. Kyung, ``Efficient neural network
  compression,'' in \emph{Proceedings of the IEEE Conference on Computer Vision
  and Pattern Recognition (CVPR)}, 2019, pp. 12\,569--12\,577.

\bibitem{cheng2017quantized}
J.~Cheng, J.~Wu, C.~Leng, Y.~Wang, and Q.~Hu, ``Quantized cnn: A unified
  approach to accelerate and compress convolutional networks,'' \emph{IEEE
  Transactions on Neural Networks and Learning Systems (TNNLS)}, vol.~29,
  no.~10, pp. 4730--4743, 2017.

\bibitem{krishnamoorthi2018quantizing}
R.~Krishnamoorthi, ``Quantizing deep convolutional networks for efficient
  inference: A whitepaper,'' \emph{arXiv preprint arXiv:1806.08342}, 2018.

\bibitem{wang2019haq}
K.~Wang, Z.~Liu, Y.~Lin, J.~Lin, and S.~Han, ``Haq: Hardware-aware automated
  quantization with mixed precision,'' in \emph{Proceedings of the IEEE
  Conference on Computer Vision and Pattern Recognition (CVPR)}, 2019, pp.
  8612--8620.

\bibitem{wang2017novel}
J.~Wang, C.~Xu, X.~Yang, and J.~M. Zurada, ``A novel pruning algorithm for
  smoothing feedforward neural networks based on group lasso method,''
  \emph{IEEE Transactions on Neural Networks and Learning Systems (TNNLS)},
  vol.~29, no.~5, pp. 2012--2024, 2017.

\bibitem{he2018amc}
Y.~He, J.~Lin, Z.~Liu, H.~Wang, L.-J. Li, and S.~Han, ``Amc: Automl for model
  compression and acceleration on mobile devices,'' in \emph{Proceedings of the
  European Conference on Computer Vision (ECCV)}, 2018, pp. 784--800.

\bibitem{chen2018constraint}
C.~Chen, F.~Tung, N.~Vedula, and G.~Mori, ``Constraint-aware deep neural
  network compression,'' in \emph{Proceedings of the European Conference on
  Computer Vision (ECCV)}, 2018, pp. 400--415.

\bibitem{frankle2019lottery}
J.~Frankle and M.~Carbin, ``The lottery ticket hypothesis: Finding sparse,
  trainable neural networks,'' in \emph{Proceedings of the International
  Conference on Learning Representations (ICLR)}, 2019.

\bibitem{lecun1990optimal}
Y.~LeCun, J.~S. Denker, and S.~A. Solla, ``Optimal brain damage,'' in
  \emph{Proceedings of the Advances in Neural Information Processing Systems
  (NeurIPS)}, 1990, pp. 598--605.

\bibitem{guo2016dynamic}
Y.~Guo, A.~Yao, and Y.~Chen, ``Dynamic network surgery for efficient dnns,'' in
  \emph{Proceedings of the Advances in Neural Information Processing Systems
  (NeurIPS)}, 2016, pp. 1379--1387.

\bibitem{ding2019global}
X.~Ding, X.~Zhou, Y.~Guo, J.~Han, J.~Liu \emph{et~al.}, ``Global sparse
  momentum sgd for pruning very deep neural networks,'' in \emph{Advances in
  Neural Information Processing Systems (NeurIPS)}, 2019, pp. 6382--6394.

\bibitem{hu2016network}
H.~Hu, R.~Peng, Y.-W. Tai, and C.-K. Tang, ``Network trimming: A data-driven
  neuron pruning approach towards efficient deep architectures,'' \emph{arXiv
  preprint arXiv:1607.03250}, 2016.

\bibitem{he2017channel}
Y.~He, X.~Zhang, and J.~Sun, ``Channel pruning for accelerating very deep
  neural networks,'' in \emph{Proceedings of the IEEE International Conference
  on Computer Vision (ICCV)}, 2017, pp. 1389--1397.

\bibitem{liu2017learning}
Z.~Liu, J.~Li, Z.~Shen, G.~Huang, S.~Yan, and C.~Zhang, ``Learning efficient
  convolutional networks through network slimming,'' in \emph{Proceedings of
  the IEEE International Conference on Computer Vision (ICCV)}, 2017, pp.
  2736--2744.

\bibitem{yu2018nisp}
R.~Yu, A.~Li, C.-F. Chen, J.-H. Lai, V.~I. Morariu, X.~Han, M.~Gao, C.-Y. Lin,
  and L.~S. Davis, ``Nisp: Pruning networks using neuron importance score
  propagation,'' in \emph{Proceedings of the IEEE Conference on Computer Vision
  and Pattern Recognition (CVPR)}, 2018, pp. 9194--9203.

\bibitem{zhao2019variational}
C.~Zhao, B.~Ni, J.~Zhang, Q.~Zhao, W.~Zhang, and Q.~Tian, ``Variational
  convolutional neural network pruning,'' in \emph{Proceedings of the IEEE
  Conference on Computer Vision and Pattern Recognition (CVPR)}, 2019, pp.
  2780--2789.

\bibitem{lin2019towards}
S.~Lin, R.~Ji, C.~Yan, B.~Zhang, L.~Cao, Q.~Ye, F.~Huang, and D.~Doermann,
  ``Towards optimal structured cnn pruning via generative adversarial
  learning,'' in \emph{Proceedings of the IEEE Conference on Computer Vision
  and Pattern Recognition (CVPR)}, 2019, pp. 2790--2799.

\bibitem{liu2019metapruning}
Z.~Liu, H.~Mu, X.~Zhang, Z.~Guo, X.~Yang, K.-T. Cheng, and J.~Sun,
  ``Metapruning: Meta learning for automatic neural network channel pruning,''
  in \emph{Proceedings of the IEEE International Conference on Computer Vision
  (ICCV)}, 2019, pp. 3296--3305.

\bibitem{lin2020hrank}
M.~Lin, R.~Ji, Y.~Wang, Y.~Zhang, B.~Zhang, Y.~Tian, and L.~Shao, ``Hrank:
  Filter pruning using high-rank feature map,'' in \emph{Proceedings of the
  IEEE Conference on Computer Vision and Pattern Recognition (CVPR)}, 2020, pp.
  1529--1538.

\bibitem{li2017pruning}
H.~Li, A.~Kadav, I.~Durdanovic, H.~Samet, and H.~P. Graf, ``Pruning filters for
  efficient convnets,'' in \emph{Proceedings of the International Conference on
  Learning Representations (ICLR)}, 2017.

\bibitem{luo2017thinet}
J.-H. Luo, J.~Wu, and W.~Lin, ``Thinet: A filter level pruning method for deep
  neural network compression,'' in \emph{Proceedings of the IEEE International
  Conference on Computer Vision (ICCV)}, 2017, pp. 5058--5066.

\bibitem{lin2019toward}
S.~Lin, R.~Ji, Y.~Li, C.~Deng, and X.~Li, ``Toward compact convnets via
  structure-sparsity regularized filter pruning,'' \emph{IEEE Transactions on
  Neural Networks and Learning Systems (TNNLS)}, vol.~31, no.~2, pp. 574--588,
  2019.

\bibitem{lemaire2019structured}
C.~Lemaire, A.~Achkar, and P.-M. Jodoin, ``Structured pruning of neural
  networks with budget-aware regularization,'' in \emph{Proceedings of the IEEE
  Conference on Computer Vision and Pattern Recognition (CVPR)}, 2019, pp.
  9108--9116.

\bibitem{lin2020channel}
M.~Lin, R.~Ji, Y.~Zhang, B.~Zhang, Y.~Wu, and Y.~Tian, ``Channel pruning via
  automatic structure search,'' in \emph{Proceedings of the International Joint
  Conference on Artificial Intelligence (IJCAI)}, 2020, pp. 673 -- 679.

\bibitem{li2015convergent}
Y.~Li, J.~Yosinski, J.~Clune, H.~Lipson, and J.~E. Hopcroft, ``Convergent
  learning: Do different neural networks learn the same representations?'' in
  \emph{Proceedings of the Advances in Neural Information Processing Systems
  (NeurIPS)}, 2015, pp. 196--212.

\bibitem{morcos2018importance}
A.~S. Morcos, D.~G. Barrett, N.~C. Rabinowitz, and M.~Botvinick, ``On the
  importance of single directions for generalization,'' in \emph{Proceedings of
  the International Conference on Learning Representations (ICLR)}, 2018.

\bibitem{zhou2018revisiting}
B.~Zhou, Y.~Sun, D.~Bau, and A.~Torralba, ``Revisiting the importance of
  individual units in cnns via ablation,'' \emph{arXiv preprint
  arXiv:1806.02891}, 2018.

\bibitem{frey2007clustering}
B.~J. Frey and D.~Dueck, ``Clustering by passing messages between data
  points,'' \emph{science}, vol. 315, no. 5814, pp. 972--976, 2007.

\bibitem{huang2018data}
Z.~Huang and N.~Wang, ``Data-driven sparse structure selection for deep neural
  networks,'' in \emph{Proceedings of the European Conference on Computer
  Vision (ECCV)}, 2018, pp. 304--320.

\bibitem{liu2019rethinking}
Z.~Liu, M.~Sun, T.~Zhou, G.~Huang, and T.~Darrell, ``Rethinking the value of
  network pruning,'' in \emph{Proceedings of the International Conference on
  Learning Representations (ICLR)}, 2019.

\bibitem{szegedy2015going}
C.~Szegedy, W.~Liu, Y.~Jia, P.~Sermanet, S.~Reed, D.~Anguelov, D.~Erhan,
  V.~Vanhoucke, and A.~Rabinovich, ``Going deeper with convolutions,'' in
  \emph{Proceedings of the IEEE Conference on Computer Vision and Pattern
  Recognition (CVPR)}, 2015, pp. 1--9.

\bibitem{he2016deep}
K.~He, X.~Zhang, S.~Ren, and J.~Sun, ``Deep residual learning for image
  recognition,'' in \emph{Proceedings of the IEEE Conference on Computer Vision
  and Pattern Recognition (CVPR)}, 2016, pp. 770--778.

\bibitem{han2015learning}
S.~Han, J.~Pool, J.~Tran, and W.~Dally, ``Learning both weights and connections
  for efficient neural network,'' in \emph{Proceedings of the Advances in
  Neural Information Processing Systems (NeurIPS)}, 2015, pp. 1135--1143.

\bibitem{aghasi2017net}
A.~Aghasi, A.~Abdi, N.~Nguyen, and J.~Romberg, ``Net-trim: Convex pruning of
  deep neural networks with performance guarantee,'' in \emph{Proceedings of
  the Advances in Neural Information Processing Systems (NeurIPS)}, 2017, pp.
  3177--3186.

\bibitem{li2006very}
P.~Li, T.~J. Hastie, and K.~W. Church, ``Very sparse random projections,'' in
  \emph{Proceedings of the 12th ACM SIGKDD international conference on
  Knowledge discovery and data mining}, 2006, pp. 287--296.

\bibitem{sulic2010efficient}
V.~Sulic, J.~Per{\v{s}}, M.~Kristan, and S.~Kovacic, ``Efficient dimensionality
  reduction using random projection,'' in \emph{15th Computer Vision Winter
  Workshop}, 2010, pp. 29--36.

\bibitem{krizhevsky2009learning}
A.~Krizhevsky, G.~Hinton \emph{et~al.}, ``Learning multiple layers of features
  from tiny images,'' 2009.

\bibitem{russakovsky2015imagenet}
O.~Russakovsky, J.~Deng, H.~Su, J.~Krause, S.~Satheesh, S.~Ma, Z.~Huang,
  A.~Karpathy, A.~Khosla, M.~Bernstein \emph{et~al.}, ``Imagenet large scale
  visual recognition challenge,'' \emph{International Journal of Computer
  Vision (IJCV)}, 2015.

\bibitem{paszke2017automatic}
A.~Paszke, S.~Gross, S.~Chintala, G.~Chanan, E.~Yang, Z.~DeVito, Z.~Lin,
  A.~Desmaison, L.~Antiga, and A.~Lerer, ``Automatic differentiation in
  pytorch,'' in \emph{Proceedings of the Advances in Neural Information
  Processing Systems (NeurIPS)}, 2017.

\bibitem{yu2019slimmable}
J.~Yu, L.~Yang, N.~Xu, J.~Yang, and T.~Huang, ``Slimmable neural networks,'' in
  \emph{Proceedings of the IEEE International Conference on Machine Learning
  (ICML)}, 2019.

\bibitem{ding2019approximated}
X.~Ding, G.~Ding, Y.~Guo, J.~Han, and C.~Yan, ``Approximated oracle filter
  pruning for destructive cnn width optimization,'' in \emph{Proceedings of the
  IEEE International Conference on Machine Learning (ICML)}, 2019.

\bibitem{luo2020autopruner}
J.-H. Luo and J.~Wu, ``Autopruner: An end-to-end trainable filter pruning
  method for efficient deep model inference,'' \emph{Pattern Recognition (PR)},
  p. 107461, 2020.

\bibitem{li2020eagleeye}
B.~Li, B.~Wu, J.~Su, G.~Wang, and L.~Lin, ``Eagleeye: Fast sub-net evaluation
  for efficient neural network pruning,'' \emph{arXiv preprint
  arXiv:2007.02491}, 2020.

\bibitem{sum1999kalman}
J.~Sum, C.-s. Leung, G.~H. Young, and W.-k. Kan, ``On the kalman filtering
  method in neural network training and pruning,'' \emph{IEEE Transactions on
  Neural Networks and Learning Systems (TNNLS)}, vol.~10, no.~1, pp. 161--166,
  1999.

\bibitem{sum1999adaptive}
J.~Sum, C.-s. Leung, G.~H. Young, L.-w. Chan, and W.-k. Kan, ``An adaptive
  bayesian pruning for neural networks in a non-stationary environment,''
  \emph{Neural Computation}, vol.~11, no.~4, pp. 965--976, 1999.

\end{thebibliography}

\begin{IEEEbiography}[{\includegraphics[width=1in,height=1.25in,clip,keepaspectratio]{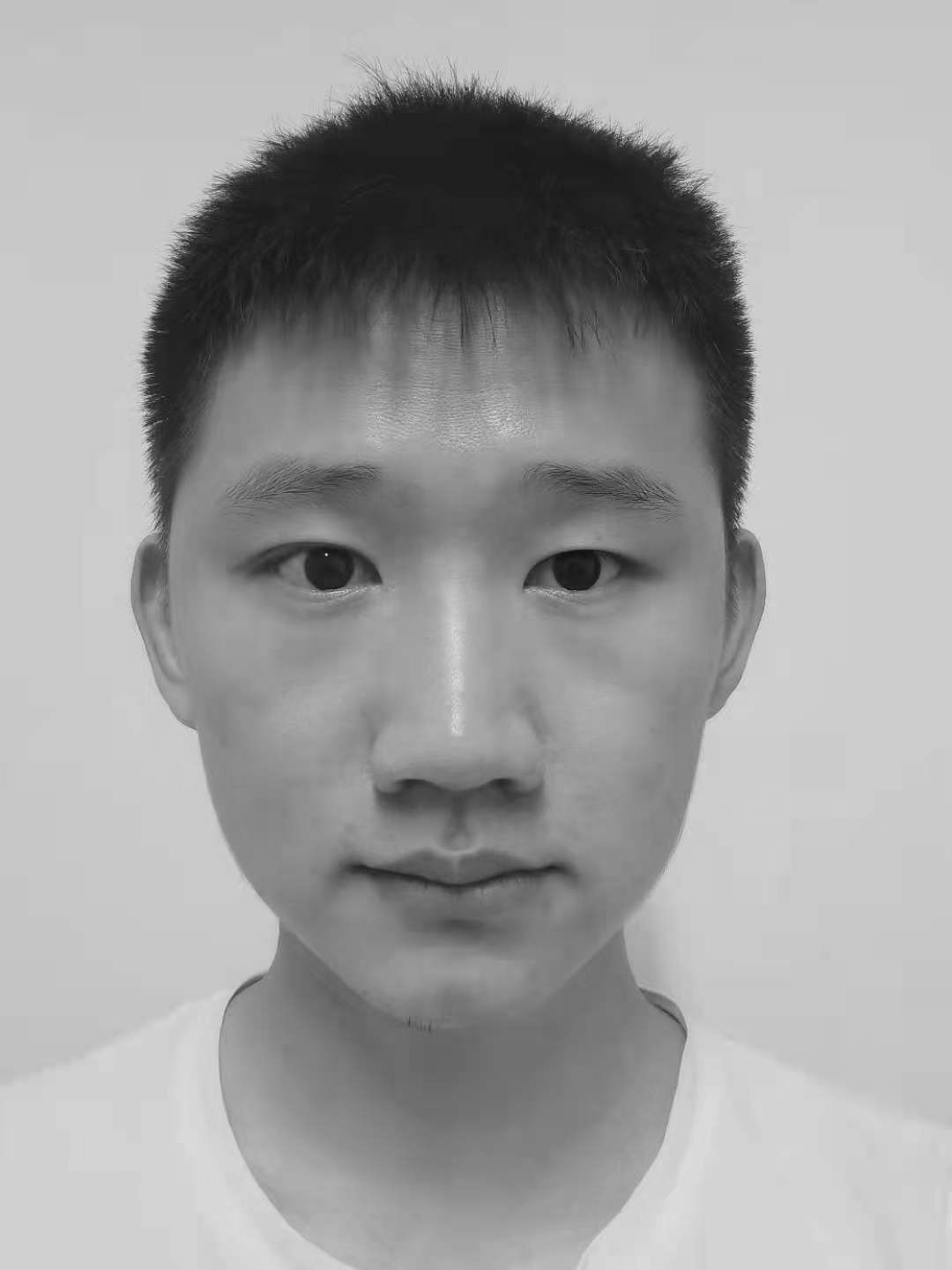}}]{Mingbao Lin}
is currently pursuing the Ph.D degree with Xiamen University, China. He has published over ten papers as the first author in international journals and conferences, including IEEE TPAMI, IJCV, IEEE TIP, IEEE TNNLS, CVPR, NeurIPS, AAAI, IJCAI, ACM MM and so on. His current research interest includes network compression \& acceleration, and information retrieval.
\end{IEEEbiography}

\begin{IEEEbiography}[{\includegraphics[width=1in,height=1.25in,clip,keepaspectratio]{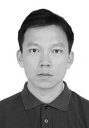}}]{Rongrong Ji}
(Senior Member, IEEE) is a Nanqiang Distinguished Professor at Xiamen University, the Deputy Director of the Office of Science and Technology at Xiamen University, and the Director of Media Analytics and Computing Lab. He was awarded as the National Science Foundation for Excellent Young Scholars (2014), the National Ten Thousand Plan for Young Top Talents (2017), and the National Science Foundation for Distinguished Young Scholars (2020). His research falls in the field of computer vision, multimedia analysis, and machine learning. He has published 50+ papers in ACM/IEEE Transactions, including TPAMI and IJCV, and 100+ full papers on top-tier conferences, such as CVPR and NeurIPS. His publications have got over 10K citations in Google Scholar. He was the recipient of the Best Paper Award of ACM Multimedia 2011. He has served as Area Chairs in top-tier conferences such as CVPR and ACM Multimedia. He is also an Advisory Member for Artificial Intelligence Construction in the Electronic Information Education Committee of the National Ministry of Education.
\end{IEEEbiography}

\begin{IEEEbiography}[{\includegraphics[width=1in,height=1.25in,clip,keepaspectratio]{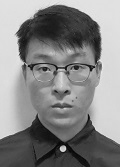}}]{Shaojie Li}
studied for his B.S. degrees in FuZhou University, China, in 2019. He is currently trying to pursue a M.S. degree in Xiamen University, China. His research interests include model compression and computer vision.
\end{IEEEbiography}

\begin{IEEEbiography}[{\includegraphics[width=1in,height=1.25in,clip,keepaspectratio]{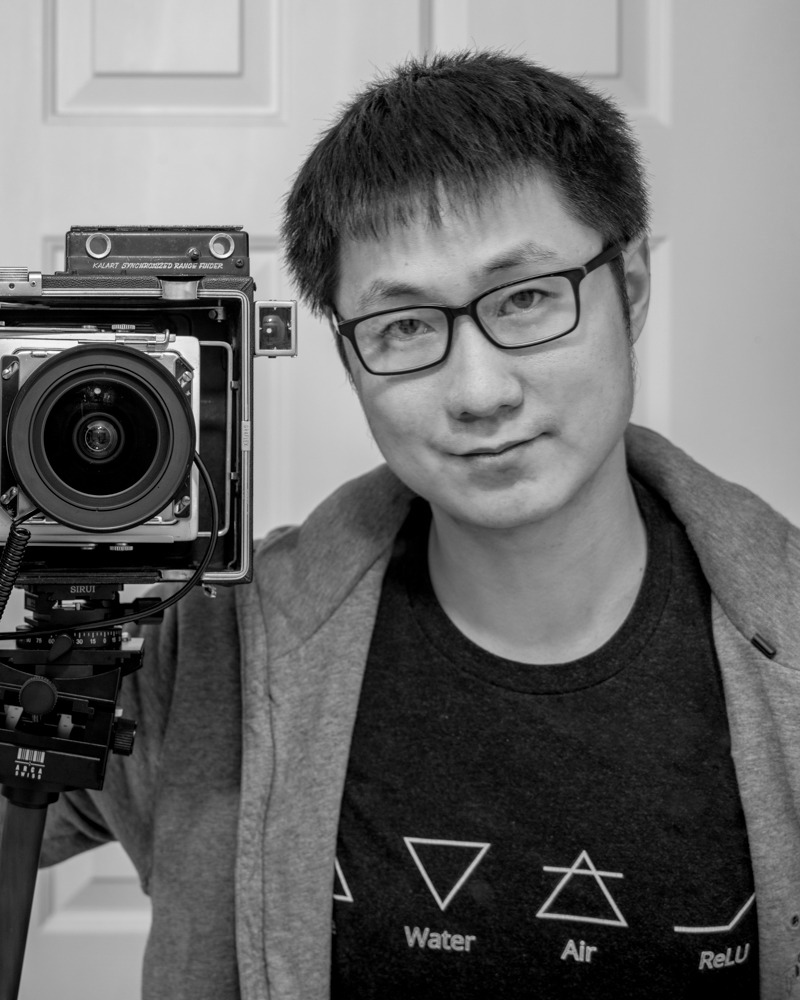}}]{Yan Wang}
works as a software engineer in Search at Pinterest. With a Ph.D degree on Electrical Engineering from Columbia University, Yan published over 20 papers on top international conferences and journals, and holds 10 US or international patents. He has broad interests on deep learning's applications on multimedia retrieval.
\end{IEEEbiography}

\begin{IEEEbiography}[{\includegraphics[width=1in,height=1.25in,clip,keepaspectratio]{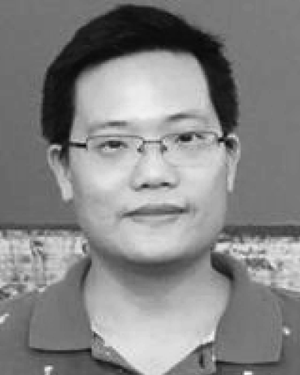}}]{Yongjian Wu} received the master degree in computer science from Wuhan University, China, in 2008. He is currently the Expert Researcher and the Director of the Youtu Lab, Tencent Co., Ltd. His research interests include face recognition, image understanding, and large scale data processing.
\end{IEEEbiography}

\begin{IEEEbiography}[{\includegraphics[width=1in,height=1.25in,clip,keepaspectratio]{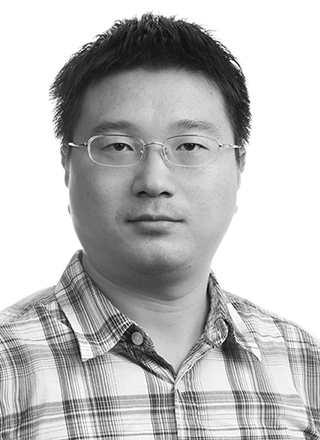}}]{Feiyue Huang} received the Ph.D degree from Tsinghua University, China, in 2008. He is currently the Director of Tencent Youtu Lab. His research interests include face recognition, pattern recognition, and machine learning.
\end{IEEEbiography}

\begin{IEEEbiography}[{\includegraphics[width=1in,height=1.25in,clip,keepaspectratio]{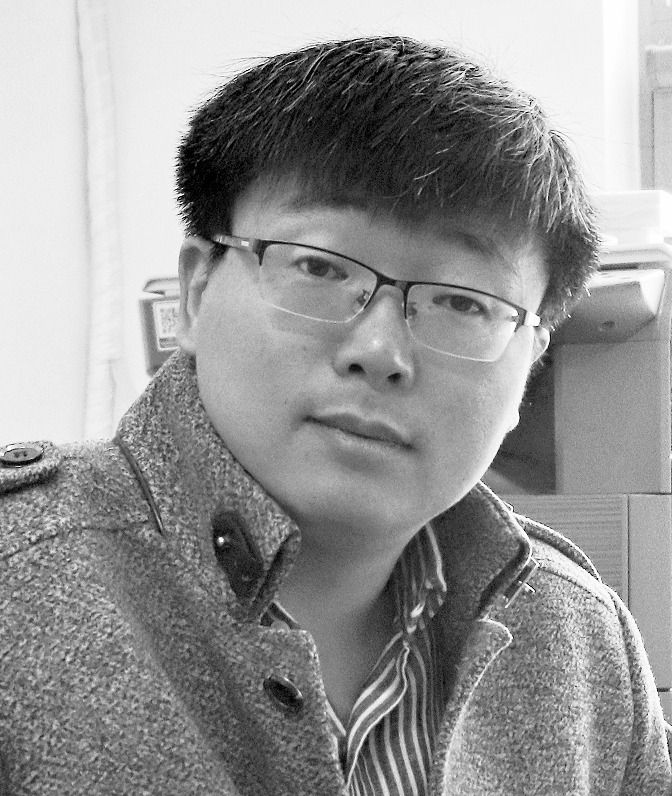}}]{Qixiang Ye} (Senior Member, IEEE) received the B.S. and M.S. degrees in mechanical and electrical engineering from Harbin Institute of Technology, China, in 1999 and 2001, respectively, and the Ph.D. degree from the Institute of Computing Technology, Chinese Academy of Sciences in 2006. He has been a professor with the University of Chinese Academy of Sciences since 2009, and was a visiting assistant professor with the Institute of Advanced Computer Studies (UMIACS), University of Maryland, College Park until 2013. His research interests include image processing, object detection and machine learning. He has published more than 100 papers in refereed conferences and journals including IEEE CVPR, ICCV, ECCV, NeurIPS, TNNLS, and PAMI.
\end{IEEEbiography}

\end{document}